\documentclass{article} 
\usepackage{iclr2026_conference,times}

\usepackage{amsmath,amsfonts,bm}

\def\eqref#1{equation~\ref{#1}}

\def\1{\bm{1}}

\DeclareMathAlphabet{\mathsfit}{\encodingdefault}{\sfdefault}{m}{sl}
\SetMathAlphabet{\mathsfit}{bold}{\encodingdefault}{\sfdefault}{bx}{n}

\usepackage[UTF8,scheme=plain,fontset=fandol]{ctex}
\usepackage{hyperref}
\usepackage{url}

\usepackage{graphicx}
\usepackage{booktabs}
\usepackage{multirow}
\usepackage{array}
\usepackage{amsmath}
\usepackage{amssymb}
\usepackage{xcolor}
\usepackage{makecell}
\usepackage{caption}
\usepackage{subcaption}
\usepackage{colortbl}

\usepackage{listings}
\usepackage{wrapfig}
\definecolor{lightgreen}{RGB}{139,195,74}
\usepackage[most]{tcolorbox}
\tcbuselibrary{breakable,skins}
\newtcolorbox{promptbox}[2][]{%
  breakable, enhanced,
  colback=#2!8, colframe=#2!75!black, boxrule=0.4pt, arc=3pt, outer arc=3pt,
  colbacktitle=#2!80!black, coltitle=white, fonttitle=\bfseries,
  title={#1},
  top=6pt, bottom=6pt, left=8pt, right=8pt,
  toptitle=4pt, bottomtitle=4pt, lefttitle=8pt, righttitle=8pt,
  before skip=10pt, after skip=10pt,
  fontupper=\small, parbox=false}
\graphicspath{{figures/}}

\newcommand{\zh}[1]{#1}

\newcommand{\benchmark}{Culture-MMT}
\newcommand{\judger}{Judger}
\newcommand{\typesym}{Symbol}
\newcommand{\typevis}{Visual}
\newcommand{\typehyb}{Hybrid}

\title{Beyond Words Alone: Disentangling Distributed Evidence for Visually Grounded Cultural Translation}
\author{
\textbf{Dingming Li}\textsuperscript{1,2,*} \quad
\textbf{Yiwen Qiu}\textsuperscript{1,2,*} \quad
\textbf{Linjuan Wu}\textsuperscript{2} \quad
\textbf{Hongxing Li}\textsuperscript{1} \\
\textbf{ Zixuan Wang}\textsuperscript{1} \quad
\textbf{Xinze Lyu}\textsuperscript{2} \quad
\textbf{Ye Guo}\textsuperscript{2} \quad
\textbf{Daoxin Zhang}\textsuperscript{2} \quad
\textbf{Zhe Xu}\textsuperscript{2} \\
\textbf{ Weiming Lu}\textsuperscript{1} \quad
\textbf{Yongliang Shen}\textsuperscript{1,$\dagger$} \\[6pt]
$^1$ Zhejiang University, $^2$ Xiaohongshu Inc. \\[4pt]
\texttt{\{lidingm,syl\}@zju.edu.cn}
}

\iclrfinalcopy

\begin{document}
\maketitle

\lhead{}

\renewcommand{\thefootnote}{\fnsymbol{footnote}}
\footnotetext[1]{~First two authors contributed equally. Conducted during the first author's internship at Xiaohongshu Inc.}
\footnotetext[2]{~Corresponding author.}
\renewcommand{\thefootnote}{\arabic{footnote}}

\vspace{-4mm}
\begin{abstract}
Social media enables cross-lingual communication, yet the meaning of user-generated content (UGC) is often distributed across text, images, and cultural knowledge, making translation challenging. Existing evaluations typically treat images as optional and assume textual information is sufficient, obscuring whether errors stem from missing visual evidence or inadequate cultural adaptation. We introduce \textbf{\benchmark{}}, a diagnostic benchmark for \textbf{distributed-evidence pragmatic reconstruction} in social-media translation. It contains 498 curated, de-identified UGC instances across 20 domains, categorized into \typesym{}, \typevis{}, and \typehyb{} by their required evidence. We also develop a human-aligned domain-specific judge, \judger{}, and a controlled evaluation framework that systematically varies the evidence available to translators. Across 17 model configurations, real images improve performance by +0.280 on \typevis{} and +0.197 on \typehyb{}, but only +0.050 on \typesym{}, while random images provide no comparable gains. Human-authored descriptions further show that jointly dependent cases remain more challenging even when visual facts are explicit. These results characterize culturally effective social-media translation as {meaning reconstruction from distributed evidence}, highlighting visual grounding and cultural-pragmatic adaptation as distinct model bottlenecks.
\end{abstract}

\vspace{-4mm}
\section{Introduction}
\label{sec:intro}

Social media has facilitated cross-lingual communication, with user-generated content (UGC) being both dynamic and challenging to translate.~\citep{wu2026beyond, qiu2026metrics} Unlike news or documents, the meaning of a UGC post rarely resides in text alone: authors often omit visually evident objects, scenes, attributes, and in-image text, while using internet memes and community-specific slang to convey attitudes.~\citep{yuan2026culture, pan2026multimodal, zhao2026beyond} Thus, the evidence required for culturally effective translation may be distributed across text, visual context, and shared cultural knowledge. For text-only systems, the challenge is not merely language understanding and generation, but the intrinsic absence of critical task-relevant information from the source text~\citep{gao2025multimodal, caglayan2019probing, fantinuoli2025vision}.

\vspace{-2mm}
\begin{figure}[ht]
\centering
\includegraphics[width=\textwidth]{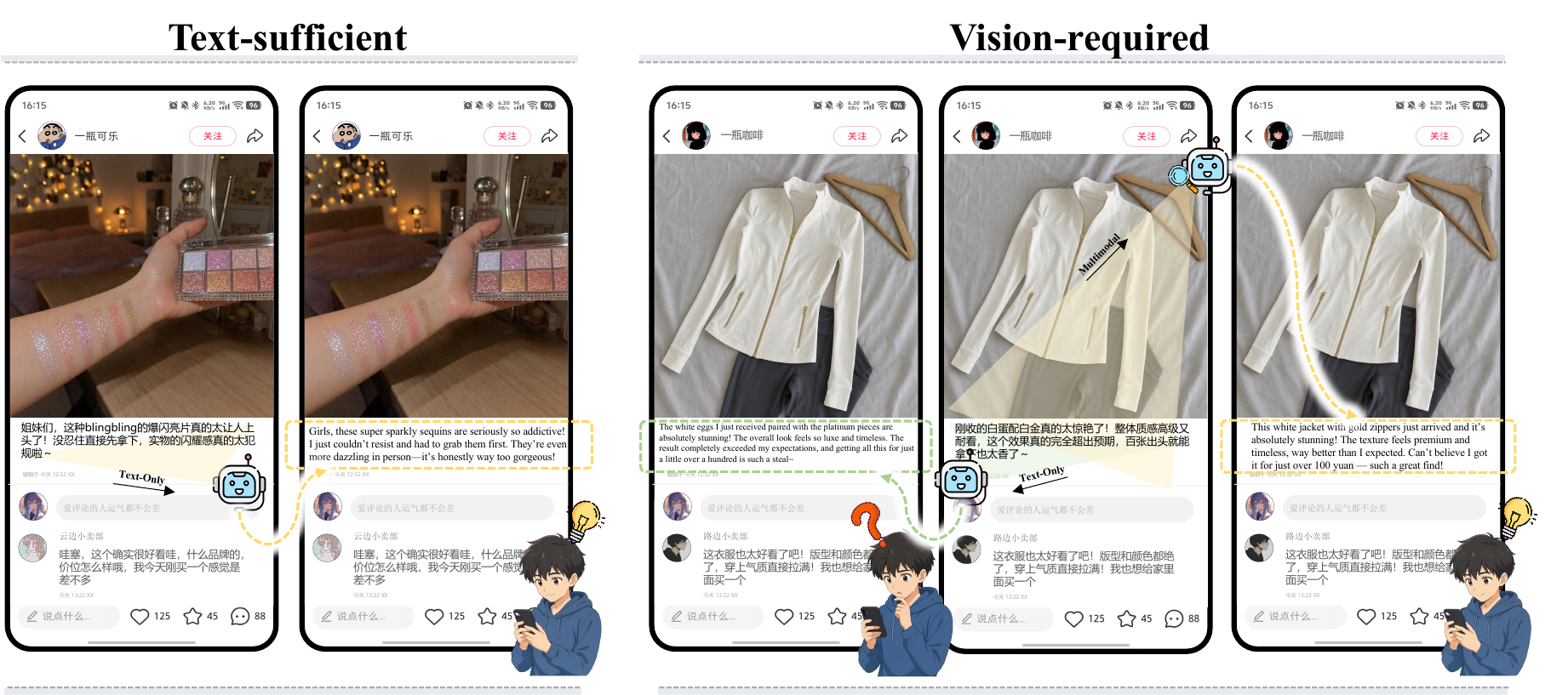}
\caption{Evidence dependence in UGC translation. Left: text-sufficient cultural slang. Right: visually dependent, multimodal input recovers missing translation evidence.}
\label{fig:cases}
\vspace{-5mm}
\end{figure}

This distributed-evidence problem remains underexplored in culturally aware translation and multimodal machine translation~\citep{villa2025cammt, pan2026multimodal}. CULTURE-MT~\citep{wu2026beyond} evaluates cultural effectiveness in UGC translation but focuses on content that can be judged from text alone. Classical multimodal translation often uses image descriptions as source text, where visual and textual information are redundant~\citep{chai2026smmbench, dash2026towards}; even ambiguity-oriented tasks typically focus on selecting the correct linguistic interpretation~\citep{villa2025cammt, gao2025multimodal}. Real-world UGC is different: images may contain entities omitted from the text, while shared cultural knowledge determines how these facts should be expressed in the target language~\citep{guo2026cross}. As shown in Figure~\ref{fig:cases}, the key question is not simply ``whether images help,'' but \textbf{whether models can access the evidence required for translation and integrate it across modalities into a culturally effective target-language expression}.

To address this gap, we introduce \benchmark{}, the first benchmark for culturally pragmatic multimodal translation from Chinese social media to English. It contains 498 de-identified real-world UGC posts across 20 domains, divided into three balanced subsets~\citep{wu2026beyond} of 166 instances based on required evidence: \typesym{} (internet cultural knowledge, without image dependence), \typevis{} (key facts conveyed by images, without cultural memes), and \typehyb{} (requiring visual fact recognition and cultural knowledge). This design isolates visual evidence completion, cultural-pragmatic mapping, and their interaction.

We further train a human-aligned domain-specific judge~\citep{wu2026beyond, belmadani2026judges, fu2025reliable}, \judger{}, and establish a controlled evaluation protocol in which the judge uses the same human-verified image description, while the translator receives one of four evidence conditions: no visual evidence, the real image, a random image, or a human-authored description. Across 17 model configurations, real images yield gains of +0.280 on \typevis{} and +0.197 on \typehyb{}, but only +0.050 on \typesym{}, while random images provide no comparable gains. Human-authored descriptions alleviate part of the perceptual gap, yet \typehyb{} remains harder than \typevis{} even when visual facts are explicit, showing that visual fact access does not imply successful cultural adaptation. Overall, \benchmark{} evaluates not only translation quality but also whether failures arise from evidence access, visual fact recovery, or cultural adaptation, revealing these as related but non-interchangeable capability stages.

Our main contributions are as follows:

\begin{itemize}
    \item We introduce \textbf{\benchmark{}} and establish a distributed-evidence perspective for social-media translation, with 498 real-world, de-identified UGC posts categorized by three evidence-dependence types.

    \item We develop a \textbf{diagnostic evaluation framework} for visually grounded cultural effectiveness, featuring a human-aligned \judger{} and a controlled protocol that fixes evaluation evidence while varying translator-side evidence.

    \item We evaluate 17 model configurations under four evidence conditions, showing that visual gains are evidence-specific and that visual fact recovery and cultural adaptation are related but non-interchangeable capabilities.
    
\end{itemize}

\section{The \benchmark{} Benchmark}
\label{sec:benchmark}

\subsection{Task Definition and Evidence-Dependency Types}
\label{sec:task}

\benchmark{} is a Chinese-to-English UGC translation diagnostic benchmark. Each sample is
a real social-media note $x = (t, b, I)$, where $t$, $b$ and $I$ denote the title, the body
and one or more images~\citep{yang2024polclip}, respectively. Unlike the default assumption that the source text is
semantically self-contained~\citep{chai2026smmbench}, we posit that the evidence needed to produce a culturally
effective translation $y$ may be distributed across text, images and shared cultural
knowledge. The evaluation compares a text-only translator $y_{txt} = f(t, b)$ with a
multimodal translator $y_{mm} = f(t, b, I)$, so as to measure how different evidence
accessibility changes cultural effectiveness.

\begin{table}[ht]
\caption{The three evidence-dependency types (166 notes each).}
\vspace{-4mm}
\label{tab:types}
\begin{center}
\small
\renewcommand{\arraystretch}{1.3}
\begin{tabular}{@{}p{1.5cm} p{5.0cm} p{5.0cm}@{}}
\toprule
\textbf{Type} & \textbf{Meaning} & \textbf{Example} \\
\midrule
\typesym{} &
Ambiguity comes from internet memes and in-group jargon; cultural knowledge alone
suffices and the image is not needed. &
``\zh{避雷}'' is a warning to steer clear of a bad product, not about lightning;
``\zh{爸爸}'' refers to a (male) idol, not a father. \\
\typevis{} &
The key sense is polysemous and can only be fixed by the image; a dictionary does not
help resolve it. &
``\zh{单反}'' with a tennis photo means a one-handed backhand,
not the default DSLR camera. \\
\typehyb{} &
Both are required at once: the jargon needs cultural knowledge \emph{and} the image
must disambiguate it. &
``\zh{肥宅复健 Day 1}'' is self-mocking jargon; only the photo says
if ``\zh{复健}'' means exercise or injury rehab. \\
\bottomrule
\end{tabular}
\end{center}
\vspace{-4mm}
\end{table}

We divide the data into three groups by ``which kind of external evidence a translator must
access to complete the translation'' (Table~\ref{tab:types}): for \typesym{}, the missing
information comes from the shared cultural knowledge behind internet slang, memes and platform
jargon, and no image is required; for \typevis{}, the key semantics are supplied by the image~\citep{zheng2025multi};
\typehyb{} depends simultaneously on visual facts and cultural knowledge. The three types are
therefore not a monotonically increasing difficulty ladder but a diagnostic grouping used to
localize the source of failure: \typesym{} probes cultural transfer, \typevis{} probes
visual-fact recovery, and \typehyb{} probes their joint requirement. Labels are assigned by
humans according to the translation mechanism; model pre-labeling is used only as auxiliary
information during filtering. Frequent visual mechanisms include scene disambiguation, proper
names and nicknames, elided reference, and in-image attributes and OCR.

\subsection{Data Construction}
\label{sec:construction}

\begin{figure}[t]
\centering
\includegraphics[width=\textwidth]{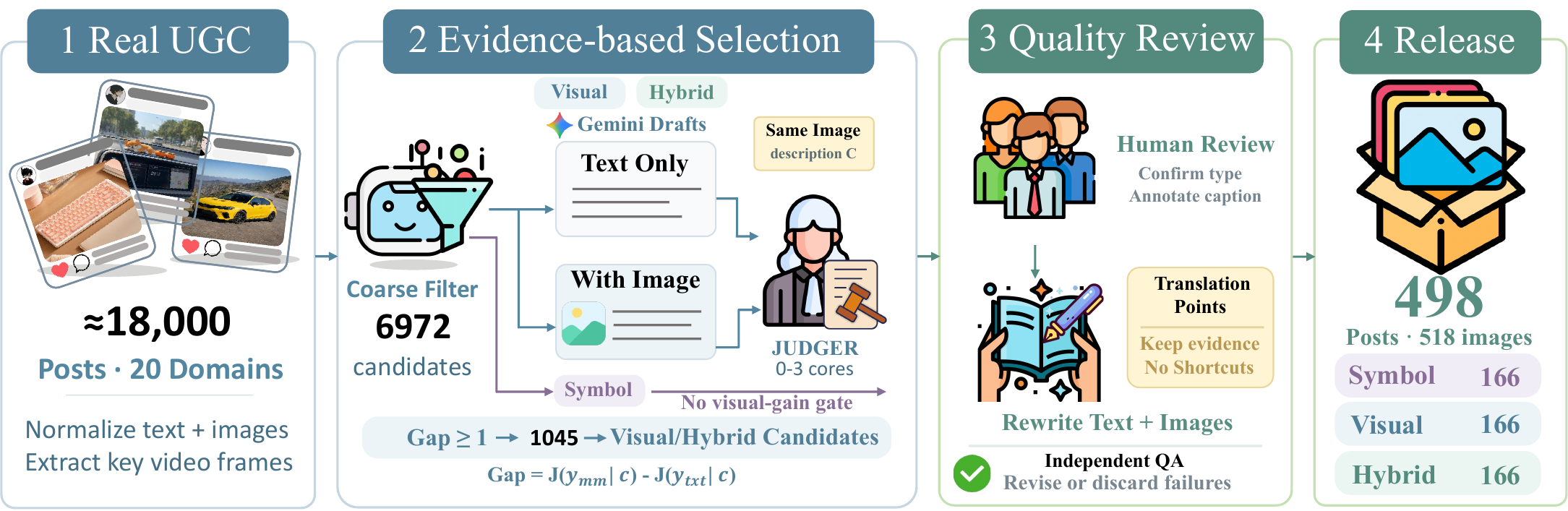}
\caption{The data-construction pipeline of \benchmark{}. Models are used for candidate enrichment; the evidence-dependency type and final inclusion are decided by humans.}
\label{fig:pipeline}
\vspace{-2mm}
\end{figure}

The data-construction pipeline is shown in Figure~\ref{fig:pipeline} and comprises candidate collection, human filtering, two-stage differential verification, privacy-preserving rewriting, and final quality control~\citep{zheng2023judging}. Candidates are collected from real Chinese social-media UGC across $20$ verticals in non-overlapping crawling rounds (about $1{,}000$ notes per round); for video notes, only static frames are retained. After normalization and corrupted-sample removal, a Gemini multimodal model~\citep{team2023gemini} generates text-only and multimodal draft translations and temporary type labels for human filtering~\citep{shao2024visual}. Three AI-background graduate students conducted filtering and privacy-preserving rewriting over two weeks, judging each note for a valid cultural or visual translation point and overriding model labels when needed. \typevis{} and \typehyb{} require key meanings unrecoverable from text alone, while \typesym{} requires a cultural translation challenge without necessary visual evidence. Details are provided in Appendix~\ref{app:annotation}.

Human judgment alone can still let ``pseudo-hard'' cases slip in---cases that strong models can translate correctly without the image~\citep{ma2025noise}. In a review of $159$ passed samples with Gemini, about one third were correctly translatable without the image, with $20$--$30\%$ noise in the previous difficulty labels. We therefore introduce a two-stage automatic filter. The first is a feature-based coarse filter without LLM calls: it keeps samples of $5$--$30$ characters containing Chinese characters and a visual or strong textual signal (e.g., code-switching, elided references, named entities), while removing catch-all phrases, generic templates, and the Easy tier. The second is a model-based fine filter: Gemini generates an image description, a text-only translation, and a multimodal translation; our text \judger{} then scores the two translations under the same visual context, yielding
\begin{equation}
\mathrm{gap} = \mathrm{score}(y_{mm} \mid c) - \mathrm{score}(y_{txt} \mid c),
\end{equation}
where $c$ is the image description. We keep samples with $\mathrm{gap} \ge 1$ and rank them by gap. This directly measures the image's contribution to translation quality, avoiding the false positives and negatives of word-overlap heuristics. Across all collection rounds, about $18{,}000$ raw candidates were processed, with $6{,}972$ surviving the coarse filter and $1{,}045$ the differential filter ($\mathrm{gap} \ge 1$). Together with human-confirmed \typesym{} controls, these candidates underwent human re-review and privacy-preserving rewriting, yielding $498$ final notes, with $166$ per dependency type.

All candidates originate from real UGC~\citep{liao2025culture}. To protect privacy, we rewrite traceable surface details while preserving the original scenario, language phenomena, evidence-dependency type, and core translation point. Specifically, we preserve the vertical, register, narrative function, and the image's role in understanding the text, while replacing identifiable people, nicknames, numbers, brands, and visual details. Titles, bodies, and images are jointly rewritten into coherent social-media notes to maintain their natural contextual flow. Images are regenerated using one of four role-specific strategies and pass perceptual-hash and structural-similarity checks; the full protocol is given in Appendix~\ref{app:construction}. Thus, \benchmark{} is grounded in real UGC while releasing privacy-preserved instances that retain the original translation challenges.

\subsection{Data Analysis}
\label{sec:analysis}

\begin{figure}[t]
\centering
\includegraphics[width=\textwidth]{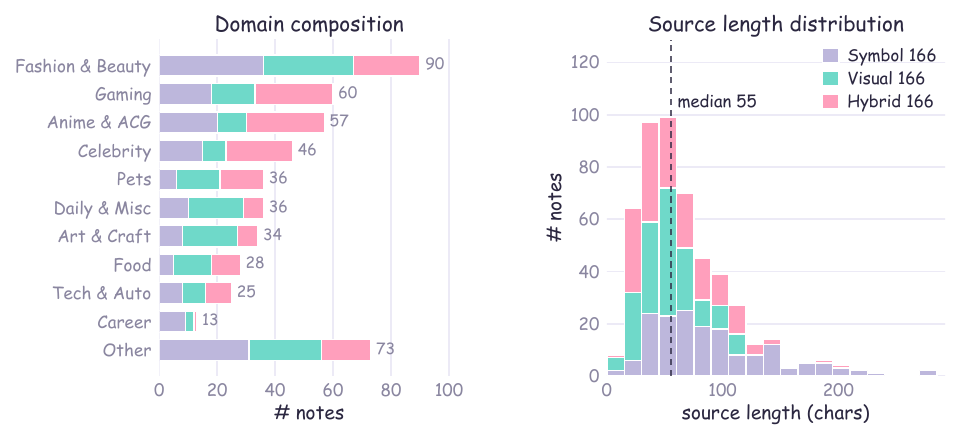}
\caption{\textbf{Dataset statistics.} (Left) Notes are balanced across the three evidence-dependency types (166 each) over 20 domains. (Right) Source length (title+body): mean 66.9, median 55, and p90 117 characters (dashed line marks the median).}
\label{fig:data}
\end{figure}

This section summarizes the statistics of the released \benchmark{} set (Figure~\ref{fig:data}). The sealed set contains $498$ real-sourced, de-identified notes with $518$ regenerated images, selected from $1{,}329$ original images by manually identifying the key image, yielding a note-to-image ratio of about $1{:}1.04$. The three evidence-dependency types are strictly balanced at $166$ each, enabling controlled cross-type comparison. Of the $498$ notes, $461$ are image-text notes and $37$ use video cover frames; $75$ are title-only notes without a body. Source texts have a median length of $55$ characters, a mean of $66.9$, and a p90 of $117$. The $20$ verticals follow a long-tailed distribution, led by fashion and beauty ($90$), gaming ($60$), anime and ACG ($57$), celebrities and fandom ($46$), and pets ($36$), with food, tech and cars, workplace, parenting, and fitness forming the long tail. Privacy-rewriting intensity is reported in Appendix~\ref{app:construction}.


\section{Evaluation Framework}
\label{sec:framework}

\subsection{Translator Evidence Conditions and the Cultural-Effectiveness Metric}
\label{sec:conditions}

Four translator input conditions form an intervention on evidence accessibility: \texttt{Text Only} provides only the text; \texttt{Multimodal} provides the native image matching the sample; \texttt{Random Image} provides an image drawn from another sample, used to distinguish relevant evidence from a generic multimodal input effect; and \texttt{Human Caption} provides a human-verified image description as evidence that does not participate in translation, used to relieve the visual-perception burden~\citep{qiu2026metrics}. Each condition keeps the translation instruction unchanged; the full prompts are given in Appendix~\ref{app:protocol}.

Cultural effectiveness is judged by the domain-specialized Chinese \judger{} trained in this work~\citep{wu2026beyond, hu2026colt}, loaded on self-hosted vLLM~\citep{kwon2023efficient} with $8$-way tensor parallelism ($\mathrm{tp}=8$), using a fixed deterministic decoding configuration on the judge side to ensure reproducible and consistent scoring. Following the rubric of the CULTURE-MT~\citep{wu2026beyond}, we report two metrics: the \textbf{mean score} is the average of the $0$--$3$ scale, and the \textbf{cultural-effectiveness rate} is the proportion of samples scoring $2$ or above ($2$--$3$, i.e., semantically and pragmatically adequate), with $0$--$1$ counted as ineffective; the visual gain $\Delta$ is the difference of the multimodal condition over the text-only condition on these metrics.

\subsection{The Controlled Evidence Protocol}
\label{sec:protocol}

The core of the protocol is to separate ``what evidence the sample contains'' from ``what evidence the translator can access.'' The \judger{} always reads the same human-verified image description for all translator conditions~\citep{liao2025culture}, because that description characterizes the sample's ground truth rather than any single translator's input; the evidence intervention occurs only on the translator side. We do not provide the judge with an enumerated ``ambiguous word $\to$ correct meaning'' answer ledger, to avoid degrading scoring into answer matching. With the judge's information fixed, score changes across conditions can be interpreted as differences in the translator's acquisition and use of evidence, providing a common reference for the subsequent failure attribution.

Moreover, we delegate ``understanding the image'' to human annotation and provide the judge with only the textual description: the benefit is that the judge's visual ability no longer forms a ceiling on the evaluation~\citep{guo2025care}, at the cost of the granularity of the human description becoming a new ceiling. The validity of \judger{} on our benchmark, including its agreement with human annotations and comparison with existing LLM judges, is evaluated in \ref{sec:judge}.

\subsection{Judge Validity}
\label{sec:judge}

\begin{wraptable}{r}{0.5\textwidth}
\vspace{-\baselineskip}
\caption{Models used for judge validation.}
\label{tab:judge_configs}
\centering
\small
\renewcommand{\arraystretch}{1.2}
\begin{tabular}{@{}l|l@{}}
\toprule
\textbf{Category} & \textbf{Model configuration} \\
\midrule
\multirow{4}{*}{Commercial (6)}
 & Gemini-3.1-pro (high) \\
 & Gemini-3.1-pro (low) \\
 & GPT-5.5,\, Kimi-K3 \\
 & GLM-5.3-Flash,\, Qwen3.7-Flash \\
\midrule
\multirow{3}{*}{Open-weight (4)}
 & Qwen3.5-397B-A17B \\
 & Qwen3.5-122B-A10B \\
 & Qwen3.5-35B-A3B,\, Qwen3.5-27B \\
\bottomrule
\end{tabular}
\vspace{-0.5\baselineskip}
\end{wraptable}

\textbf{Validation design.} The \judger{} is built on a Qwen3.5-27B~\citep{qwen35blog} base and aligned to the $0$--$3$ human standard of cultural effectiveness through two-stage supervised fine-tuning~\citep{kim2026cpc} and GRPO~\citep{shao2024deepseekmath, guo2025deepseek}; the full training configuration is given in Appendices~\ref{app:protocol} and~\ref{app:implementation}. We sample $20$ notes from each of the three evidence-dependency types, and the $10$ model configurations listed in Table~\ref{tab:judge_configs} each generate translations under text-only and multimodal conditions, yielding $1{,}200$ candidates across the two modes. After removing missing or incompletely annotated pairs, we keep $1{,}160$ human--\judger{} valid pairs covering $58$ notes. Under a binary split, the agreement between the
\judger{} and human judgment is $0.820$, Cohen's $\kappa$ is $0.459$, and the effective-class F1 is $0.886$; the four-class weighted $\kappa$ is $0.455$. Its recall on the effective class is higher than on the ineffective class ($0.89$ vs.\ $0.56$), indicating that the \judger{} can serve as an automatic proxy for cultural effectiveness overall. The full sampling and annotation procedure is given in Appendix~\ref{app:annotation}.

Based on the human judge results, we balance $249$ effective and $289$ ineffective samples, and further randomly sample to construct a $498$-sample class-balanced set for comparing the domain-trained judge with general LLM judges. Table~\ref{tab:judge} shows that the \judger{} achieves the highest accuracy ($0.781$) and Cohen's $\kappa$ ($0.562$) on~\citep{cohen1960coefficient} this balanced set. The raw Qwen3.5-27B of the same base is already close to the \judger{} on the binary split, but domain training markedly improves the four-tier calibration on which the main experiment relies: four-class accuracy rises from $0.398$ to $0.522$, $\kappa$ from $0.225$ to $0.342$, and mean absolute error falls from $0.745$ to $0.586$.

\begin{table}[t]
\caption{\textbf{Evaluator agreement with human experts on the balanced test set.} -01 and -23 denote the invalid (scores 0--1) and valid (scores 2--3) classes, respectively.}
\label{tab:judge}
\begin{center}
\renewcommand{\arraystretch}{1.15}
\resizebox{\textwidth}{!}{%
\begin{tabular}{l cc cccc cc}
\toprule
\textbf{Judge} & \textbf{ACC} & \textbf{Cohen's $\kappa$}
& \textbf{P-01} & \textbf{R-01} & \textbf{F1-01}
& \textbf{P-23} & \textbf{R-23} & \textbf{F1-23} \\
\midrule
\textbf{\judger{} (ours)} & \textbf{0.781} & \textbf{0.562} & \textbf{1.000} & \textbf{0.562} & \textbf{0.720} & \textbf{0.696} & \textbf{1.000} & \textbf{0.820} \\
Gemini-3.1-pro           & 0.767 & 0.534 & 0.809 & 0.699 & 0.750 & 0.735 & 0.835 & 0.782 \\
Qwen3.5-397B             & 0.733 & 0.466 & 0.736 & 0.727 & 0.731 & 0.730 & 0.739 & 0.735 \\
Qwen3-32B                & 0.552 & 0.104 & 1.000 & 0.104 & 0.189 & 0.528 & 1.000 & 0.691 \\
Qwen3.5-27B              & 0.773 & 0.546 & 0.809 & 0.715 & 0.759 & 0.745 & 0.831 & 0.786 \\
\bottomrule
\end{tabular}%
}
\end{center}
\vspace{-4mm}
\end{table}

\section{Experiments}
\label{sec:experiments}

\subsection{Experimental Setup}
\label{sec:setup}

All experiments are conducted on 498 samples in \benchmark{}. The main results compare two translator input conditions, Text Only and Multimodal, while the judge uses only human caption; each model configuration is evaluated under both text-only and image-text conditions, with the judge always receiving the same human-verified visual evidence. We evaluate 17 model configurations in total, including the proprietary models GPT-5.5/5.6-sol~\citep{singh2025openai}, Kimi-K3~\citep{team2026kimi}, Gemini-3.1-pro (high/low)~\citep{team2023gemini}, and Qwen3.7-Flash~\citep{yang2025qwen3}, as well as the open-source models GLM-5.3-Flash~\citep{glm5team2026glm5vibecodingagentic}, Qwen3.5-397B-A17B~\citep{qwen35blog}, Qwen3.5-122B-A10B~\citep{qwen35blog}, Qwen3.5-35B-A3B~\citep{qwen35blog}, Qwen3.5-27B/9B/4B~\citep{qwen35blog}, Qwen3-VL-235B/32B/8B~\citep{yang2025qwen3} and Gemma-4-31B~\citep{team2026gemma}. Access details and inference parameters are given in Appendix~\ref{app:implementation}.

\textbf{Metrics.} The primary metrics are the mean cultural-effectiveness score and effective rate (the proportion scoring $2$--$3$), aligned with \textsc{Culture-MT} for cross-comparability. Unless otherwise stated, visual gain $\Delta$ is the multimodal minus text-only difference in mean $0$--$3$ score, reported for \typesym{}, \typevis{}, and \typehyb{}; overall result is their weighted aggregate.

\subsection{Main Results}
\label{sec:main}

\begin{table}[t]
\caption{\textbf{Cultural effectiveness and score distribution across 17 models.} Reporting rates (\%), score breakdowns, and multimodal gains ($\Delta$). \textbf{Bold} marks group best.}
\label{tab:main}
\begin{center}
\resizebox{\textwidth}{!}{%
\renewcommand{\arraystretch}{1.1}
\begin{tabular}{l cccccc cccccc cc}
\toprule
& \multicolumn{6}{c}{\textbf{Text-only}} & \multicolumn{6}{c}{\textbf{Multimodal}} & & \\
\cmidrule(lr){2-7}\cmidrule(lr){8-13}
\textbf{Model} & Valid & Inval. & 0 & 1 & 2 & 3 & Valid & Inval. & 0 & 1 & 2 & 3
& \textbf{$\Delta$Valid} & \textbf{$\Delta$Mean} \\
\midrule
\rowcolor{gray!15}
\multicolumn{15}{l}{\textbf{Proprietary Models}} \\
Gemini-3.1-Pro (high) & 78.1 & 21.9 & 3.8 & 18.1 & 20.9 & \textbf{57.2} & \textbf{82.7} & \textbf{17.3} & 2.2 & \textbf{15.1} & 20.9 & 61.9 & +4.62 & +0.108 \\
Gemini-3.1-Pro (low)  & \textbf{79.1} & \textbf{20.9} & \textbf{3.6} & \textbf{17.3} & \textbf{22.9} & 56.2 & \textbf{82.7} & \textbf{17.3} & \textbf{1.6} & 15.7 & \textbf{24.1} & 58.6 & +3.61 & +0.080 \\
GPT-5.6-Sol               & 72.1 & 27.9 & 5.2 & 22.7 & 16.9 & 55.2 & 81.1 & 18.9 & 2.2 & 16.7 & 16.9 & \textbf{64.3} & \textbf{+9.03} & \textbf{+0.211} \\
GPT-5.5               & 73.9 & 26.1 & 3.8 & 22.3 & 20.5 & 53.4 & 79.1 & 20.9 & 2.4 & 18.5 & 21.1 & 58.0 & +5.22 & +0.112 \\
Qwen3.7-flash         & 72.9 & 27.1 & 4.4 & 22.7 & 20.7 & 52.2 & 80.3 & 19.7 & 3.6 & 16.1 & 23.3 & 57.0 & +7.43 & +0.131 \\
\midrule
\rowcolor{gray!15}
\multicolumn{15}{l}{\textbf{Open-source models}} \\
Kimi-K3               & \textbf{75.9} & \textbf{24.1} & \textbf{4.0} & \textbf{20.1} & 20.9 & \textbf{55.0} & \textbf{80.3} & \textbf{19.7} & \textbf{2.4} & 17.3 & 20.3 & \textbf{60.0} & +4.42 & +0.110 \\
GLM-5.3-Flash         & 72.1 & 27.9 & 5.2 & 22.7 & 22.1 & 50.0 & 78.1 & 21.9 & 3.8 & 18.1 & 20.5 & 57.6 & +6.02 & +0.151 \\
Qwen3.5-397B-A17B     & 65.3 & 34.7 & 7.2 & 27.5 & \textbf{27.9} & 37.4 & 74.7 & 25.3 & 4.4 & 20.9 & 27.1 & 47.6 & +9.44 & +0.225 \\
Qwen3-VL-235B         & 62.8 & 37.2 & 14.7 & 22.5 & 19.1 & 43.7 & 72.4 & 27.6 & 11.7 & \textbf{15.9} & 20.3 & 52.1 & +9.65 & +0.211 \\
Qwen3.5-122B-A10B     & 60.8 & 39.2 & 6.6 & 32.5 & 27.3 & 33.5 & 68.1 & 31.9 & 3.8 & 28.1 & 29.7 & 38.4 & +7.23 & +0.149 \\
Qwen3.5-35B-A3B       & 53.4 & 46.6 & 9.0 & 37.5 & 25.3 & 28.1 & 60.0 & 40.0 & 5.0 & 34.9 & 27.9 & 32.1 & +6.63 & +0.147 \\
Qwen3-VL-32B          & 52.2 & 47.8 & 14.7 & 33.1 & 25.9 & 26.3 & 62.6 & 37.4 & 10.0 & 27.3 & 26.3 & 36.4 & +10.44 & +0.251 \\
Gemma-4-31B           & 66.3 & 33.7 & 6.2 & 27.5 & 26.5 & 39.8 & 72.5 & 27.5 & 4.2 & 23.3 & 22.7 & 49.8 & +6.22 & +0.183 \\
Qwen3.5-27B           & 62.5 & 37.5 & 7.2 & 30.3 & 27.3 & 35.1 & 73.9 & 26.1 & 4.4 & 21.7 & 26.3 & 47.6 & +11.45 & \textbf{+0.267} \\
Qwen3.5-9B            & 47.6 & 52.4 & 12.1 & 40.4 & 27.1 & 20.5 & 57.8 & 42.2 & 8.2 & 33.9 & 28.9 & 28.9 & +10.24 & +0.225 \\
Qwen3-VL-8B           & 42.4 & 57.6 & 21.9 & 35.7 & 26.1 & 16.3 & 51.4 & 48.6 & 18.1 & 30.5 & 28.3 & 23.1 & +9.04 & +0.197 \\
Qwen3.5-4B            & 39.6 & 60.4 & 12.8 & 47.6 & 26.9 & 12.7 & 52.4 & 47.6 & 11.4 & 36.1 & \textbf{31.1} & 21.3 & \textbf{+12.85} & +0.229 \\
\bottomrule
\end{tabular}%
}
\end{center}
\vspace{-4mm}
\end{table}

\textbf{Once matching evidence is accessible, overall quality rises consistently.} In
Table~\ref{tab:main}, all $17$ configurations improve in mean score ($+0.080$ to $+0.267$, mean $+0.176$), the cultural-effectiveness rate increases by $7.86$ percentage points on average, and the score distribution shifts noticeably toward $3$. The trend spans model families and scales, showing that text input systematically omits information in UGC that can be supplied by the image; but this only proves that vision is useful, not that models exploit evidence matched to the sample.

\begin{figure}[t]
\centering
\includegraphics[width=\textwidth]{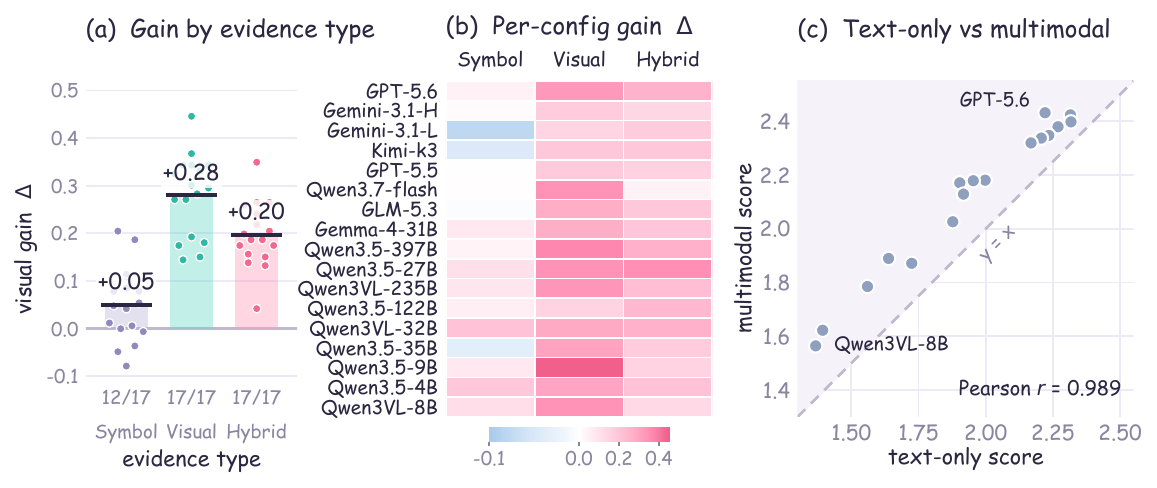}
\vspace{-7mm}
\caption{\textbf{Visual gains across evidence types and configurations.} \textbf{(a)} Mean gain $\Delta$: \typesym{} remains unchanged, while \typevis{} and \typehyb{} gain significantly. \textbf{(b)} Per-configuration breakdown. \textbf{(c)} Text-only vs. multimodal scores ($r=0.989$).}
\label{fig:gains}
\end{figure}

\begin{table}[t]
\caption{\textbf{Visual gains across 17 models.} Reporting mean scores under Text-only and Multimodal inputs, and their differences ($\Delta = \text{Multimodal} - \text{Text-only}$) overall and per dependency type.}
\vspace{-4mm}
\label{tab:gain}
\begin{center}
\renewcommand{\arraystretch}{1.12}
\resizebox{\textwidth}{!}{%
\begin{tabular}{l ccc ccc}
\toprule
\textbf{Model} & \textbf{Text-only} & \textbf{Multimodal} & \textbf{$\Delta$ Overall}
& \textbf{$\Delta$ \typesym{}} & \textbf{$\Delta$ \typevis{}} & \textbf{$\Delta$ \typehyb{}} \\
\midrule
\rowcolor{gray!15}
\multicolumn{7}{l}{\textbf{Proprietary Models}} \\
Gemini-3.1-Pro (high) & 2.315 & 2.424 & +0.108 & +0.012 & +0.175 & +0.139 \\
Gemini-3.1-Pro (low)  & 2.317 & 2.398 & +0.080 & $-$0.078 & +0.145 & +0.175 \\
GPT-5.6-Sol               & 2.221 & 2.432 & +0.211 & +0.048 & +0.325 & +0.259 \\
GPT-5.5               & 2.235 & 2.347 & +0.112 & +0.000 & +0.181 & +0.157 \\
Qwen3.7-Flash         & 2.207 & 2.337 & +0.131 & +0.006 & +0.343 & +0.042 \\
\midrule
\rowcolor{gray!15}
\multicolumn{7}{l}{\textbf{Open-source models}} \\
Kimi-K3               & 2.269 & 2.380 & +0.110 & $-$0.048 & +0.193 & +0.187 \\
GLM-5.3-Flash         & 2.169 & 2.319 & +0.151 & $-$0.006 & +0.271 & +0.187 \\
Qwen3.5-397B-A17B     & 1.954 & 2.179 & +0.225 & +0.042 & +0.367 & +0.265 \\
Qwen3-VL-235B         & 1.917 & 2.129 & +0.211 & +0.085 & +0.331 & +0.218 \\
Qwen3.5-122B-A10B     & 1.877 & 2.026 & +0.149 & +0.054 & +0.151 & +0.241 \\
Qwen3.5-35B-A3B       & 1.725 & 1.871 & +0.147 & $-$0.036 & +0.301 & +0.175 \\
Qwen3-VL-32B          & 1.639 & 1.890 & +0.251 & +0.205 & +0.283 & +0.265 \\
Gemma-4-31B           & 1.998 & 2.181 & +0.183 & +0.078 & +0.271 & +0.199 \\
Qwen3.5-27B           & 1.904 & 2.171 & +0.267 & +0.108 & +0.343 & +0.349 \\
Qwen3.5-9B            & 1.560 & 1.785 & +0.225 & +0.078 & +0.446 & +0.151 \\
Qwen3-VL-8B           & 1.367 & 1.564 & +0.197 & +0.115 & +0.343 & +0.133 \\
Qwen3.5-4B            & 1.394 & 1.623 & +0.229 & +0.187 & +0.295 & +0.205 \\
\midrule
\textbf{Mean (17 configs)} & \textbf{1.945} & \textbf{2.121} & \textbf{+0.176} & \textbf{+0.050} & \textbf{+0.280} & \textbf{+0.197} \\
\bottomrule
\end{tabular}%
}
\end{center}
\vspace{-3mm}
\end{table}

\textbf{The visual benefit is evidence-selective, not a generic multimodal dividend.} Figure~\ref{fig:gains} and Table~\ref{tab:gain} show that across 17 configurations, the average gains on \typesym{}, \typevis{}, and \typehyb{} are $+0.050$, $+0.280$, and $+0.197$, respectively. While \typevis{} and \typehyb{} show positive gains for every configuration, \typesym{} remains small and unstable, with 5 configurations showing no gain or a drop. Images thus help consistently only when they provide information missing from the text; when translation relies mainly on shared cultural knowledge, they are often unhelpful and may distract. The lower gain on \typehyb{} further confirms this: jointly dependent samples retain symbolic difficulty that images cannot resolve, which persists in \ref{sec:caption} even with human descriptions.

\textbf{Visual access alleviates the information bottleneck but does not close the capability gap.} Text-only and multimodal scores are highly consistent across models (Pearson $r=0.989$, computed from Table~\ref{tab:gain}), with strong models remaining ahead after receiving images. Images mainly restore missing sample-level facts, while final quality remains constrained by language generation, pragmatic inference, and cultural transfer. Visual evidence thus alleviate input-side information constraints but cannot guarantee quality translation.

\textbf{Scaling raises absolute quality, but evidence use does not scale monotonically.} For the Qwen3.5 dense series (4B/9B/27B), the multimodal mean rises from $1.623$ to $1.785$ to $2.171$, while gains are $+0.229/+0.225/+0.267$; for Qwen3-VL (8B/32B/235B), the mean rises from $1.564$ to $1.890$ to $2.129$, with gains of $+0.197/+0.251/+0.211$. Native visual pre-training is not decisive: Qwen3.5-27B, a general-purpose model, achieves the highest gain ($+0.267$), while Qwen3-VL does not lead in $\Delta$. Scaling therefore raises the performance ceiling but does not guarantee better image use, making evidence use an evaluation dimension independent of scale and modality architecture.

\textbf{Absolute quality and visual gain reveal different model bottlenecks.} Gemini-3.1-pro (high) achieves $2.424$ multimodal with only $+0.108$ gain; Qwen3.5-27B scores lower ($2.171$) but achieves the highest gain ($+0.267$); GPT-5.6 achieves both the highest multimodal score ($2.432$) and a large gain ($+0.211$). Lower baselines tend to yield larger $\Delta$ (see Table~\ref{tab:gain}), so a large $\Delta$ may reflect effective image use or simply greater room for improvement. Thus, the multimodal score measures translation quality, while $\Delta$ measures how much vision closes the information gap, jointly distinguishing translation quality from visual reliance.

\subsection{Evidence-Attribution Analysis}
\label{sec:attribution}

\subsubsection{Random-Image Control}
\label{sec:random}

Because the type differences in the main results could still stem from an attention or prompting effect of an ``extra modality,'' we set up a \texttt{Random Image} condition that provides the translator with an image from another sample while keeping the prompt, model and judge unchanged, to test whether the gain requires the evidence to match the current sample.

\textbf{Mismatched evidence cannot substitute for missing evidence.} On $4$ representative configurations, the average change of random images over text-only is \typesym{} $-0.074$, \typevis{} $-0.005$, \typehyb{} $-0.077$ (overall $-0.052$), whereas the same models with real images improve \typevis{} and \typehyb{} by $+0.297$ and $+0.232$. The main result is thus not a ``one extra input modality'' effect: models do read the image, but only visual evidence matching the current text can reconstruct the omitted contextual meaning, while a mismatched image is at best unhelpful and even introduces noise that distracts generation (Table~\ref{tab:ablation})---in sharp contrast to the visual-redundancy phenomenon of classic MMT, where randomly swapping the image is almost lossless.

\begin{table}[t]
\caption{\textbf{Evidence-attribution ablations across four representative configurations.} Mean scores ($0$--$3$) across dependency types under four translator inputs: \emph{Text-only}, \emph{Random image} (placebo control), \emph{Human caption} (oracle perception), and \emph{Real multimodal}.}
\label{tab:ablation}
\begin{center}
\small
\renewcommand{\arraystretch}{1.12}
\begin{tabular}{l l cccc}
\toprule
\textbf{Model configuration} & \textbf{Condition} & \textbf{\typesym{}} & \textbf{\typevis{}} & \textbf{\typehyb{}} & \textbf{Overall} \\
\midrule
\multirow{4}{*}{Gemini-3.1-pro (high)}
 & Text-only       & 2.530 & 2.277 & 2.139 & 2.315 \\
 & Random image    & 2.458 & 2.307 & 2.115 & 2.293 \\
 & Human caption   & 2.512 & 2.440 & 2.283 & 2.412 \\
 & Real multimodal & 2.542 & 2.452 & 2.277 & 2.424 \\
\midrule
\multirow{4}{*}{Qwen3.5-397B-A17B}
 & Text-only       & 2.217 & 1.879 & 1.765 & 1.954 \\
 & Random image    & 2.120 & 1.873 & 1.560 & 1.851 \\
 & Human caption   & 2.259 & 2.307 & 2.006 & 2.191 \\
 & Real multimodal & 2.259 & 2.247 & 2.030 & 2.179 \\
\midrule
\multirow{4}{*}{Qwen3.5-27B}
 & Text-only       & 2.060 & 1.916 & 1.735 & 1.904 \\
 & Random image    & 2.024 & 1.904 & 1.657 & 1.861 \\
 & Human caption   & 2.151 & 2.355 & 2.084 & 2.197 \\
 & Real multimodal & 2.169 & 2.259 & 2.084 & 2.171 \\
\midrule
\multirow{4}{*}{Qwen3.5-35B-A3B}
 & Text-only       & 1.952 & 1.711 & 1.512 & 1.725 \\
 & Random image    & 1.861 & 1.681 & 1.512 & 1.685 \\
 & Human caption   & 2.084 & 2.163 & 1.765 & 2.004 \\
 & Real multimodal & 1.916 & 2.012 & 1.687 & 1.871 \\
\midrule
\multirow{4}{*}{\textbf{Mean (4 configs)}}
 & \textbf{Text-only}       & \textbf{2.190} & \textbf{1.946} & \textbf{1.788} & \textbf{1.974} \\
 & \textbf{Random image}    & \textbf{2.116} & \textbf{1.941} & \textbf{1.711} & \textbf{1.923} \\
 & \textbf{Human caption}   & \textbf{2.251} & \textbf{2.316} & \textbf{2.035} & \textbf{2.201} \\
 & \textbf{Real multimodal} & \textbf{2.221} & \textbf{2.242} & \textbf{2.020} & \textbf{2.161} \\
\bottomrule
\end{tabular}
\end{center}
\vspace{-4mm}
\end{table}

\subsubsection{The Form of Visual Evidence}
\label{sec:caption}

The real-image condition requires a model to both recognize visual facts and use them for translation, and the two failures are hard to separate. The \texttt{Human Caption} condition directly provides human-verified visual facts, bypassing perception; comparing it with \texttt{Text Only} and \texttt{Multimodal} roughly distinguishes whether failure comes from ``visual facts not recovered'' or ``facts obtained but target-language transfer not completed.'' The judge input is kept identical across the three conditions.

\textbf{Having visual facts available does not equal completing cultural transfer.} On 4 representative configurations, the overall mean under \texttt{Human Caption} ($2.201$) is slightly higher than direct image reading ($2.161$), as is the score on \typevis{} ($2.316$ vs.\ $2.242$), suggesting that models may miss some task-relevant visual facts, but the gap is limited. Meanwhile, \typehyb{} under \texttt{Human Caption} reaches only $2.035$, below \typevis{}'s $2.316$. Since the two groups are not strictly matched, this gap is not a pure causal estimate of cultural ability, but it shows that even with visual facts textualized, jointly dependent samples retain additional difficulty beyond what images can provide (Table~\ref{tab:ablation}). 

Together, the three interventions suggest three failure stages: inaccessible evidence, insufficient visual-fact recovery, and insufficient cultural-pragmatic transfer. Representative qualitative cases localizing errors to each bottleneck stage are illustrated in Appendix~\ref{app:failure_cases}.

\section{Conclusion}
\label{sec:conclusion}

We move beyond the conventional view of whether images improve translation and argue that culturally effective social-media translation is fundamentally \textbf{meaning reconstruction from distributed evidence}. \benchmark{} uses three evidence-dependence types and controlled interventions to diagnose model failures across three stages: inaccessible evidence, insufficient fact recovery, and insufficient cultural adaptation. Experiments across 17 models show that visual perception and cultural knowledge are not interchangeable: the former recovers omitted facts, while the latter enables pragmatic adaptation. This perspective provides standardized evaluation and diagnostic tools for studying complex pragmatic representation in next-generation multimodal models.

\bibliography{iclr2026_conference}
\bibliographystyle{iclr2026_conference}

\appendix
\section{Related Work}
\label{sec:related}

\subsection{Multimodal Machine Translation}

Multimodal machine translation (MMT) leverages visual context to mitigate textual ambiguity~\citep{feng2025multimodal,chen2025make,gao2025multimodal,liu2025detect,hatami2025leveraging}. However, standard benchmarks assume semantically self-contained text, leading prior work to question the necessity of vision~\citep{futeral2025towards,kumar2025does,xiong2025giift}. While recent studies incorporate cultural awareness and visual grounding~\citep{yao2024benchmarking,yuan2026culture,villa2025cammt,satar2025seeing,zhou2025hanfu}, they conflate multimodal perception with cultural understanding. In contrast, \benchmark{} reframes UGC translation as \textbf{meaning reconstruction from distributed evidence}. Crucially, visual fact recovery and cultural adaptation represent distinct, non-interchangeable bottlenecks: accessing visual facts does not guarantee pragmatic adequacy. Through evidence-dependency categorization and controlled interventions, \benchmark{} systematically disentangles model failures across evidence access, visual grounding, and cultural transfer.

\subsection{Culture-Aware and UGC Translation}

Recent studies investigate cultural awareness in text-only translation~\citep{yao2024benchmarking,yuan2026culture,liao2025culture,guo2025care} and extend it to multimodal contexts through visual grounding and meme transcreation~\citep{villa2025cammt,pan2026multimodal,zhao2026beyond,zheng2025multi}. In user-generated content (UGC), however, prior work largely focuses on colloquial noise or multimodal dialogue~\citep{michel2018mtnt,liang2022msctd}, while cultural benchmarks like \textsc{Culture-MT}~\citep{wu2026beyond,qiu2026metrics} assume the source text is semantically self-contained. In contrast, we extend cultural evaluation to visually dependent UGC where essential meaning is omitted from text. By introducing \typesym{} as a text-only control and applying controlled evidence conditions, \benchmark{} systematically disentangles visual fact recovery from cultural adaptation.

\section{Limitations}
\label{sec:limitations}

\noindent\textbf{Language and Cultural Scope.}
\benchmark{} focuses exclusively on the Chinese-to-English direction. Although the theoretical framework of distributed evidence is generalizable to other high-context online communication, Chinese social media exhibits highly specific cultural memes and elliptical patterns. Validating whether the observed evidence-utilization trends generalize to other typologically distinct language pairs and cultural spaces remains future work.

\noindent\textbf{Sample Granularity across Domains and Mechanisms.}
With 498 curated notes (166 per evidence cell), our benchmark provides sufficient rigor for macro-level comparisons across the three dependency types. However, its statistical power is limited for finer-grained sub-analyses across the 20 verticals or visual grounding mechanisms (e.g., OCR text, scene disambiguation, and cultural metaphors).

\noindent\textbf{Evaluation Ceiling of Textual Proxies.}
To prevent perceptual errors and hallucinations of multimodal judges from confounding evaluation, our \textsc{Judger} evaluates translations against human-verified textual descriptions rather than raw images. While this protocol ensures reproducible and controlled scoring across conditions, it inherently shifts the evaluation ceiling to the descriptive granularity of human annotations---subtle visual nuances omitted in the captions cannot be credited.

\section{Model Failure-Mode Cases}
\label{app:failure_cases}

\begin{figure}[t]
\centering
\includegraphics[width=\textwidth]{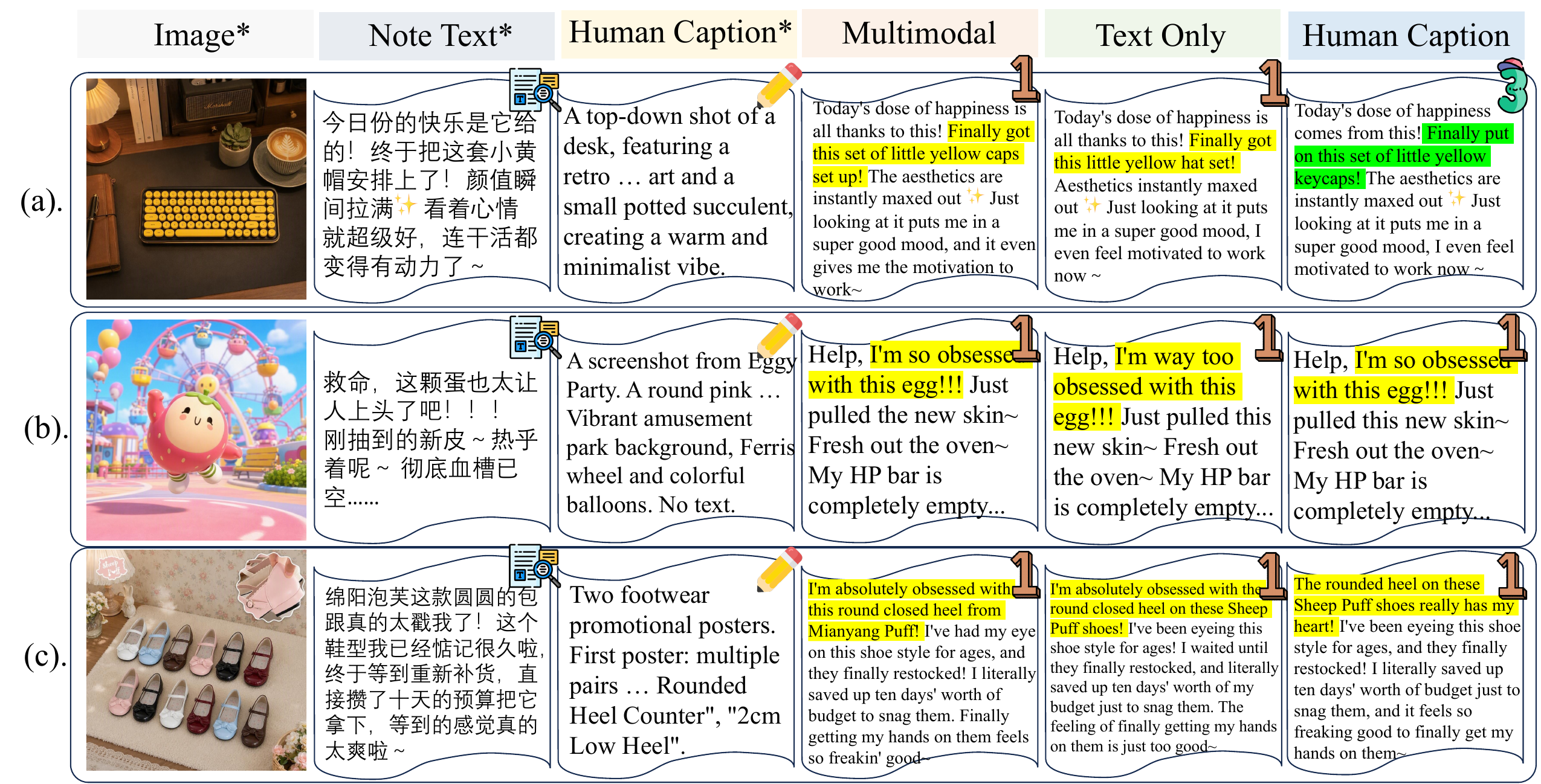}
\caption{Three representative failure modes of multimodal cultural translation: \textbf{(a)}
visual-evidence recognition, \textbf{(b)} cross-modal evidence integration, and \textbf{(c)}
cultural transfer and target-language expression. Columns marked ``*'' are benchmark information;
unmarked columns show Gemini-3.1-pro outputs under different translator input conditions, with the upper-right number indicating the score.}
\label{fig:failure}
\vspace{-4mm}
\end{figure}

Figure~\ref{fig:failure} presents three representative failure modes in multimodal cultural translation: visual-evidence recognition, cross-modal evidence integration, and cultural transfer with target-language expression. Each panel juxtaposes the source image-text, model translations under different evidence conditions, and an error note, allowing the bottleneck to be localized to a specific stage.

\textbf{Failure-mode descriptions.} \textbf{(a) Visual-evidence recognition failure} occurs when the model fails to recover a key visual fact needed for translation. Here, the multimodal model renders ``小黄帽'' as the ambiguous ``little yellow caps'', while the correct expression is ``little yellow keycaps''. The human-caption condition resolves the error, indicating that the bottleneck is failure to recognize the objects as keyboard keycaps.

\textbf{(b) Cross-modal evidence-integration failure} occurs when the model recognizes the visual fact but fails to align it with an elliptical or referential expression in the text. Here, ``这颗蛋'' remains ``this egg'', whereas the intended meaning is ``this strawberry Eggy character''. The human-caption condition does not resolve the error, suggesting that the visual fact is available but is not correctly integrated with the textual reference.

\textbf{(c) Cultural-transfer and target-language-expression failure} occurs when the model recovers the visual fact and source meaning but fails to express them naturally in the target language. Here, the rounded heel of the shoes is translated as the awkward ``round closed heel'', whereas ``rounded heel counter'' is more precise and natural. This indicates a bottleneck in target-language terminology and cultural-pragmatic expression.
\section{Evaluation Protocol and Full Prompts}
\label{app:protocol}

\subsection{Why the Judge Needs Visual Evidence}
\label{app:why_visual}

We assess cultural effectiveness with the domain-specialized \judger{} we train (on aQwen3.5-27B~\citep{qwen35blog} base). Unlike a generic LLM-as-a-judge~\citep{fu2025reliable}, this task additionally requires the judge to
determine whether a translation recovers information carried by the image but not made explicit in the text. If the judge saw only the source text, it might misjudge a correct visual completion as an unfounded addition; and if different translator conditions corresponded to different judge evidence, the change of evaluation condition would be confounded. We therefore fix the judge input
across all translator conditions: the \judger{} always receives the same human-verified image description, fixing the judge's basis on the sample's attributes so that between-group differences correspond to the evidence the translator can access. The agreement of the \judger{} with human scoring and the comparison results are given in \ref{sec:judge}.

\subsection{Translator Input Conditions and the Full Prompt}
\label{app:translator_prompt}

The four translator conditions share the same translation instruction and change only the accessible visual evidence. \texttt{Text Only} inputs only the title and body to be translated, with no visual information attached; \texttt{Multimodal} inputs the real image of the current note through the model's native image part; \texttt{Random Image} inputs an unrelated image from another sample; \texttt{Human Caption} places the human-verified Chinese image description before
the content to be translated and explicitly marks it as auxiliary evidence that need not be translated. Apart from the form of the evidence, the prompt structure and output requirements are identical across conditions. The full translator prompt used in the experiments is given below.

\begin{promptbox}[Translator prompt (shared across all conditions)]{gray}
You are a multilingual translation expert specialized in translating Chinese user-generated social media content into English. Please translate the following [CONTENT] from Chinese into English.

Important instructions:

1. Placeholders in the form of \#占位x\# must not be translated. Preserve each placeholder exactly and keep it in the corresponding position in the translated text.

2. Output only the translated result. Do not include any additional content, notes, or explanations.

3. Preserve all structural tags such as \textless{}title\textgreater{}\textless{}/title\textgreater{}, \textless{}content\textgreater{}\textless{}/content\textgreater{}, etc., and output the translation using the same format.

4. The source text is user-generated social media content. The translation should:\\
\hspace*{1.05em}- Faithfully preserve the original emotional tone (e.g., excitement, sarcasm, complaint, humor, irony).\\
\hspace*{1.05em}- Sound natural and idiomatic to native English speakers on social platforms.\\
\hspace*{1.05em}- Avoid overly formal, academic, or literal phrasing.

5. When the original text contains culture-specific expressions, slang, or implicit context that may not be immediately clear to English readers:\\
\hspace*{1.05em}- Translate them into an equivalent expression that conveys the same intent and emotion.\\
\hspace*{1.05em}- Do not add explanations or annotations.

6. Do not introduce information unsupported by the source post. Information grounded in the accompanying images is part of the source post and may be made explicit when necessary for a faithful translation.\\
\hspace*{1.05em}Do not omit emotionally or pragmatically important details or structure.

The [CONTENT] to be translated is as follows:\\
\{content\}

Please provide the translated result:
\end{promptbox}

Under the \texttt{Human Caption} condition, the translator first sees an image description
explicitly marked ``do NOT translate,'' and then reads the title and body to be translated. The
input form is as follows:

\begin{promptbox}[Additional input under the \texttt{Human Caption} condition]{gray}
{}[Image information for disambiguation, do NOT translate this part]\\
{}[Description of the accompanying picture] \textless{}human-annotated caption\textgreater{}

\textless{}title\textgreater{}…\textless{}/title\textgreater{}\\
\textless{}content\textgreater{}…\textless{}/content\textgreater{}
\end{promptbox}

\subsection{Judge Training, Input Organization, and Full Prompt}
\label{app:judge_prompt}

\textbf{\judger{} training.} The \judger{} is built on a Qwen3.5-27B base. We collect source text from public social-media platforms and use multiple translation models to construct about $300$K source--translation pairs covering different quality levels, of which about $200$K are labeled by Gemini and $100$K by humans. Training has three stages: first, supervised fine-tuning
on $150$K samples balanced across the $0$--$3$ scale drawn from the Gemini-labeled pool; then continued fine-tuning on $40$K balanced human-labeled samples; finally GRPO on $20$K samples drawn from the human-labeled pool, further aligning the judge with human evaluation. At inference it is deployed via vLLM on $8$ $\times$ $80$\,GB NVIDIA H800 GPUs with $8$-way tensor parallelism ($\mathrm{tp}=8$).

The \judger{} rates a translation on a $0$--$3$ scale for semantic accuracy, linguistic naturalness and cultural fit: $0$ means serious errors with core semantics not conveyed; $1$ means the main idea is recognizable but there are clear mistranslations, omissions or expression problems; $2$ means the main information is accurate and grammar is largely correct, but there is
slight unnaturalness, cultural infelicity or stylistic weakening; $3$ means accurate semantics, natural expression and fit to the English social-media context~\citep{wu2026beyond}. Beyond the overall score, the
prompt specifies format constraints, eight error types, three severity levels, a structured output format and three scoring examples. For image-dependent samples, the judge reads the human-verified Chinese image description, the Chinese source and the English translation simultaneously, and uses the image description to judge whether the translation recovers visual information not explicit in
the text. The image description states only task-relevant visual facts and contains no answer to the translation point; all translator conditions use exactly the same judge input, keeping the scoring basis consistent.

The full judge prompt is given below; its input contains, in order, the human-verified image description, the Chinese source and the English translation to be evaluated.

\begin{promptbox}[Judge prompt (original Chinese, used in all experiments)]{gray}
你是一个精通中英双语、熟悉网络文化且极为严厉的翻译质量评估专家。你的任务是根据给定的上下文、原文和译文，按照以下严格的规则和格式进行评估。\\
1. 评分要求：\\
语义准确性核心要求：\\
\hspace*{1.40em}* 正确理解原文语义及情感：译文需准确反映原文的字面意思和隐含情感（如讽刺、兴奋、沮丧）。\\
\hspace*{1.40em}* 文化适应性称呼翻译：对"姐妹"、"老师"、"宝子"等称呼的翻译需考虑文化背景和社交礼仪，不能存在冒犯或歧义。\\
\hspace*{1.40em}* 符合目标文化语境：译文需符合英语文化语境和目标受众（社交媒体用户）的阅读习惯，避免中式英语。\\
\hspace*{1.40em}* 文化负载词处理：对承载特定文化、网络或语境含义的词汇（如"绝绝子"、"种草"），直译无法理解的，必须进行意译或文化替换，不可直译。\\
\hspace*{1.40em}* 单位换算准确性：涉及中式单位（如"亩"、"斤"、"里"）与国际单位的换算时，需准确并明确限定范围，避免歧义；对于中文习惯省略的单位，译文中需补充完整。\\
\hspace*{1.40em}* 专有名词准确性：对著名人名、地名、品牌名、作品名等特殊名词，要求翻译准确或使用公认译名。\\
\hspace*{1.40em}* 图文依赖判定：本条【上下文】中的『配图内容(人工精标caption)』是人工对图片内容的精标中文描述（任务导向、可能有损，并非原图本身）。评判时：(1)若译文未能体现只有借助图片才能得到的正确义项（把仅凭图片可知的意思翻错、漏译或泛化成通用词），按语义偏差从严扣分；(2)译文正确补出图中承载的信息属于应得的正确翻译，不得因『中文原文字面没写』而反扣分。\\
格式要求（一旦违反立即判0分）：\\
\hspace*{1.40em}*  \#占位x\# 是不能翻译的，否则直接0分\\
\hspace*{1.40em}*  翻译前后\textless{}comment\textgreater{}\textless{}/comment\textgreater{} \textless{}title\textgreater{}\textless{}/title\textgreater{} \textless{}content\textgreater{}\textless{}/content\textgreater{}等标签无法对应，这意味着翻译结果无法正确被提取，违反这条直接0分

2. 评分标准：\\
你需要为整篇翻译给出一个 0到3分 的总体分数，定义如下：\\
* 0分：译文有严重错误，无法传递原文语义或内容丢失/曲解。\\
* 1分：译文存在明显问题，但主要语义尚可被艰难理解。存在关键错误或文化误译，严重影响阅读体验。\\
* 2分：译文准确传达了原文的主要信息，语法基本正确，但存在表达不地道、轻微文化不适或文体功能弱化等问题。\\
* 3分：译文精准、表达自然、符合英语文化语境，语法规范，传达原文的所有信息，符合英语社交平台受众的阅读习惯。

3. 错误类型：\\
* 专有名词（著名专有名词，非著名专有名词），\\
* 特殊名词（固定称谓，特殊词汇），\\
* 语义偏差（标题主语缺失，符号遗漏/错误，微弱语法错误，文本增加/遗漏，语气不自然），\\
* 直译（直译，句子结构不地道，用词不准），\\
* 上下文缺失（原文有异议/不明确），\\
* 单位（单位增加，单位换算），\\
* 长文本（长文本专有名词，长文本翻译），\\
* 其他（中英夹杂，表情符号，政治，安全）

4. 严重程度：\\
* 严重：表示扭曲句意\\
* 中等：表示影响语义，但仍可被理解\\
* 一般：表示有错误，但语义影响不大

5. 输出格式：\\
对整句的翻译意见：xxx\\
问题1：xxx\\
对问题1的评论：xxx\\
问题1类型：（一级标签）-（二级标签）\\
问题1严重程度：xxx\\
问题2：xxx\\
对问题2的评论：xxx\\
问题2类型：（一级标签）-（二级标签）\\
问题2严重程度：xxx\\
最终分数：（只有数字）

重要格式规则说明：\\
* 如果翻译完美，则只输出`对整句的翻译意见：无问题可指出`和`最终分数：3`。\\
* 如果有一个或多个问题，必须在`对整句的翻译意见`后，按顺序列出所有问题。\\
* `对整句的翻译意见`字段不能为空，需要一个总结性地说明。

6. 评估示例：\\
* 示例1：\\
可参考的上下文：0.第一次看鸟片好紧张\\
1.\#手养鹦鹉[话题]\# \#玄凤鹦鹉[话题]\# \#合法饲养[话题]\# 尊滴好尴尬\\
2.昨天我家的狗子被不认识的大狼狗给那啥了，那狗比我家狗足足大了2倍，看到的时候已经屁股对着屁股了，我老公说这时候不能去动他们，不然会出不来，受伤的，我家狗2条后腿都悬空着的[惊恐R]\\
原文：\textless{}comment3\textgreater{}必须引产，生不出来，太危险\textless{}/comment3\textgreater{}\\
译文：\textless{}comment3\textgreater{}An abortion is necessary, it can't be born, it's too dangerous.\textless{}/comment3\textgreater{}

输出：\\
对整句的翻译意见：译文存在医学概念翻译的错误，导致语义误解。\\
问题1：An abortion is necessary\\
对问题1的评论："引产"是指人工诱发分娩以使胎儿存活，应译为"induced labor"或"an induction is necessary"。而"abortion"指终止妊娠，两者含义相反，此处为严重误译。\\
问题1类型：直译-用词不准\\
问题1严重程度：严重\\
最终分数：0

* 示例2：\\
可参考的上下文：无\\
原文：\#OPPOFindN5 手感、续航、性能、防水、闪充、AI体验全面升级。\\
更多惊喜，2 月 20 日 19:00 发布会见！\\
译文：\#OPPOFindN5 The feel, battery life, performance, water resistance, fast charging, and AI experience are all getting a major upgrade. And there's more to come! See you at the launch on February 20th at 7:00 PM!

对整句的翻译意见：无问题可指出\\
最终分数：3

* 示例3:\\
可参考的上下文：无\\
原文：\textless{}title\textgreater{}姐妹们快冲！这款面膜真的绝了！\textless{}/title\textgreater{}\\
译文：\textless{}title\textgreater{}Sisters, go quickly! This facial mask is really absolute.\textless{}/title\textgreater{}

输出：\\
对整句的翻译意见：译文生硬直译，严重削弱了原文的号召力和情感色彩。\\
问题1：go quickly! This facial mask is really absolute.\\
对问题1的评论："快冲！"在此处是强烈的购买建议，应译为"Run, don't walk!"或"You gotta get this!"。"绝了"表示极好，应译为"amazing"、"incredible"或"the best"。当前译法完全不符合英语种草文案的表达习惯。\\
问题1类型：直译-用词不准\\
问题1严重程度：中等\\
问题2：Sisters\\
对问题2的评论：在非特定语境下，"姐妹们"作为一种亲切呼语，译为"Sisters"略显生硬。根据上下文，可简化为"Guys"或"Everyone"，或直接省略。\\
问题2类型：特殊名词-特殊词汇\\
问题2严重程度：一般\\
最终分数：1

请你根据上述提示，对这个翻译内容以规定的格式进行评估，仅输出评估内容，不要输出其他无关内容。

以下是你要评估的内容： \{content\}

输出：
\end{promptbox}

The Chinese prompt above is the exact prompt used in our experiments. For readers who do not
read Chinese, the following is a faithful English translation of the same prompt (the model is
always run with the Chinese version above).

\begin{promptbox}[Judge prompt (English translation, for reference only)]{gray}
You are a translation-quality assessment expert who is fluent in both Chinese and English, familiar with internet culture, and extremely strict. Your task is to evaluate the given context, source text and translation according to the following strict rules and format.

1. Scoring requirements:\\
Core requirements for semantic accuracy:\\
\hspace*{1.40em}* Correctly understand the source semantics and emotion: the translation must accurately reflect the literal meaning and the implicit emotion of the source (e.g., sarcasm, excitement, frustration).\\
\hspace*{1.40em}* Culturally appropriate translation of address terms: translating terms of address such as "sister", "teacher", "bao-zi (dear)" must consider cultural background and social etiquette, and must not be offensive or ambiguous.\\
\hspace*{1.40em}* Fit to the target cultural context: the translation must fit the English cultural context and the reading habits of the target audience (social-media users), avoiding Chinglish.\\
\hspace*{1.40em}* Handling of culture-loaded words: for words carrying specific cultural, internet or contextual meaning (e.g., "jue-jue-zi (amazing)", "zhong-cao (recommend/plant a desire)"), when a literal translation is not understandable, paraphrase or cultural substitution is mandatory; literal translation is not allowed.\\
\hspace*{1.40em}* Accuracy of unit conversion: when Chinese units (e.g., "mu", "jin", "li") are involved, convert to international units accurately and with a clearly bounded range to avoid ambiguity; for units that Chinese habitually omits, complete them in the translation.\\
\hspace*{1.40em}* Accuracy of proper nouns: for special nouns such as famous person names, place names, brand names and work titles, the translation must be accurate or use the accepted rendering.\\
\hspace*{1.40em}* Image-text dependency judgment: the "image content (human fine-labeled caption)" in the\\
\hspace*{2.10em}[Context] is a human, task-oriented, possibly lossy Chinese description of the image (not the original image itself). When judging: (i) if the translation fails to reflect the correct sense obtainable only via the image (mistranslating, omitting, or over-generalizing into a generic word a meaning that is knowable only from the image), deduct strictly as a semantic deviation; (ii) the translation correctly supplying information carried by the image counts as a deserved correct translation and must not be penalized on the grounds that "the Chinese source does not state it literally".\\
Format requirements (any violation immediately scores 0):\\
\hspace*{1.40em}*  \#placeholder-x\# must not be translated; otherwise score 0 directly.\\
\hspace*{1.40em}*  If tags such as \textless{}comment\textgreater{}\textless{}/comment\textgreater{} \textless{}title\textgreater{}\textless{}/title\textgreater{} \textless{}content\textgreater{}\textless{}/content\textgreater{} before and after translation cannot be matched, the translation result cannot be correctly extracted; violating this scores 0 directly.

2. Scoring scale:\\
You must give one overall score from 0 to 3 for the whole translation, defined as follows:\\
* 0: the translation has serious errors and fails to convey the source semantics, or content is lost/distorted.\\
* 1: the translation has obvious problems, but the main semantics can be understood with difficulty. There are key errors or cultural mistranslations that seriously harm the reading experience.\\
* 2: the translation accurately conveys the main information of the source and grammar is largely correct, but there is unidiomatic expression, slight cultural inappropriateness, or weakened stylistic function.\\
* 3: the translation is precise, natural, fits the English cultural context, is grammatically correct, conveys all information of the source, and matches the reading habits of English social-platform audiences.

3. Error types:\\
* Proper nouns (famous proper nouns, non-famous proper nouns),\\
* Special nouns (fixed address terms, special vocabulary),\\
* Semantic deviation (missing title subject, symbol omission/error, minor grammatical error, text addition/omission, unnatural tone),\\
* Literal translation (literal translation, unidiomatic sentence structure, inaccurate wording),\\
* Missing context (source is disputable/unclear),\\
* Units (unit addition, unit conversion),\\
* Long text (long-text proper nouns, long-text translation),\\
* Other (Chinese-English code-mixing, emoji, politics, safety)

4. Severity levels:\\
* Serious: distorts the sentence meaning.\\
* Medium: affects the semantics but can still be understood.\\
* Minor: there is an error, but the semantic impact is small.

5. Output format:\\
Translation comment on the whole sentence: xxx\\
Problem 1: xxx\\
Comment on problem 1: xxx\\
Problem 1 type: (level-1 tag)-(level-2 tag)\\
Problem 1 severity: xxx\\
Problem 2: xxx\\
Comment on problem 2: xxx\\
Problem 2 type: (level-1 tag)-(level-2 tag)\\
Problem 2 severity: xxx\\
Final score: (number only)

Important formatting rules:\\
* If the translation is perfect, output only `Translation comment on the whole sentence: no problem to point out` and `Final score: 3`.\\
* If there are one or more problems, after `Translation comment on the whole sentence` list all problems in order.\\
* The `Translation comment on the whole sentence` field must not be empty; it needs a summarizing statement.

6. Scoring examples:\\
* Example 1:\\
Reference context: 0. So nervous watching a bird video for the first time\\
1. \#hand-raised parrot[topic]\# \#cockatiel[topic]\# \#legal keeping[topic]\# so embarrassing\\
2. Yesterday my dog was, you know, mounted by a big unfamiliar wolfdog, a dog a full 2x bigger than mine; by the time I saw it they were already rear-to-rear. My husband said you can't pull them apart at this moment or they'll get hurt and won't come loose; both of my dog's hind legs were dangling in the air [terrified R]\\
Source: \textless{}comment3\textgreater{}必须引产，生不出来，太危险\textless{}/comment3\textgreater{}\\
Translation: \textless{}comment3\textgreater{}An abortion is necessary, it can't be born, it's too dangerous.\textless{}/comment3\textgreater{}

Output:\\
Translation comment on the whole sentence: the translation has a medical-concept error that leads to a semantic misunderstanding.\\
Problem 1: An abortion is necessary\\
Comment on problem 1: "引产" means inducing labor artificially so that the fetus is delivered alive, and should be translated as "induced labor" or "an induction is necessary". "abortion" means terminating the pregnancy; the two have opposite meanings, so this is a serious mistranslation here.\\
Problem 1 type: Literal translation-inaccurate wording\\
Problem 1 severity: Serious\\
Final score: 0

* Example 2:\\
Reference context: none\\
Source: \#OPPOFindN5 Feel, battery life, performance, waterproofing, fast charging and the AI experience are all comprehensively upgraded. More surprises---see you at the Feb 20, 19:00 launch!\\
Translation: \#OPPOFindN5 The feel, battery life, performance, water resistance, fast charging, and AI experience are all getting a major upgrade. And there's more to come! See you at the launch on\\
February 20th at 7:00 PM!

Translation comment on the whole sentence: no problem to point out\\
Final score: 3

* Example 3:\\
Reference context: none\\
Source: \textless{}title\textgreater{}姐妹们快冲！这款面膜真的绝了！\textless{}/title\textgreater{}\\
Translation: \textless{}title\textgreater{}Sisters, go quickly! This facial mask is really absolute.\textless{}/title\textgreater{}

Output:\\
Translation comment on the whole sentence: the translation is stiff and literal, seriously weakening the source's call-to-action and emotional color.\\
Problem 1: go quickly! This facial mask is really absolute.\\
Comment on problem 1: "快冲！" here is a strong buying suggestion and should be translated as "Run, don't walk!" or "You gotta get this!". "绝了" means excellent and should be translated as "amazing", "incredible" or "the best". The current rendering does not match the expression habits of English recommendation copy.\\
Problem 1 type: Literal translation-inaccurate wording\\
Problem 1 severity: Medium\\
Problem 2: Sisters\\
Comment on problem 2: in a non-specific context, "姐妹们" as an affectionate form of address is a bit stiff when rendered as "Sisters". Depending on context, it can be simplified to "Guys" or "Everyone", or dropped entirely.\\
Problem 2 type: Special nouns-special vocabulary\\
Problem 2 severity: Minor\\
Final score: 1

Following the above instructions, evaluate this translation in the specified format; output only the evaluation content and nothing else.

The content to be evaluated is as follows:\\
\{content\}

Output:
\end{promptbox}

\section{Benchmark Construction and Privacy Protection}
\label{app:construction}

\subsection{Real-UGC Candidate Collection and Normalization}
\label{app:collection}

The candidates of \benchmark{} come from real Chinese social-media UGC, collected in multiple non-overlapping time windows. We normalize the title, body, image and vertical fields, keep the static frames of video notes that can carry translation evidence, and remove samples with missing images, corrupted files or unusable text. A multimodal model generates an image description, text-only and multimodal draft translations, and a temporary evidence-dependency and vertical
label for each candidate to improve filtering efficiency; these results are auxiliary only, the final type is decided by humans, and the drafts do not enter the released data. After candidate filtering, differential verification, privacy-preserving rewriting and sealing-stage quality control, the released set contains $498$ notes covering $20$ verticals.

\subsection{Candidate Filtering and Differential Verification}
\label{app:filtering}

The filtering stage aims to obtain the three evidence-dependency types separately. The first stage is a feature-based coarse filter with no LLM call: keeping only samples of length $5$--$30$ characters containing Chinese characters, removing catch-all phrases (e.g., ``thanks / so cute /
amazing'') and templated content and discarding the Easy tier, then keeping, by regular-expression features (code-switching, elided reference, named entities, visual signals), candidates with a visual dependency or a strong textual signal, and filtering out pure cultural-symbol noise. The second stage performs differential verification on \typevis{} and \typehyb{} candidates: a Gemini
multimodal model generates an image text description, a text-only translation and a multimodal translation, and our text \judger{} (a $0$--$3$ scale, the same scoring rubric as the main experiment's judge) scores the two translations under the same image-description context; when the score difference of the multimodal translation over the text-only translation reaches the threshold ($\mathrm{gap} \ge 1$), the candidate enters human re-review, sorted by descending gap.
\typesym{} candidates are not retained by visual gain, but are confirmed by humans to require internet-culture knowledge while not depending on the image. As one measured round: $2{,}000$ candidates leave $1{,}485$ after the first stage, $312$ after the second stage ($\mathrm{gap} \ge 1$), and $75$ after human re-review; the final label of each round is decided by humans, and automatic filtering is used only to improve candidate-enrichment efficiency. The input convention of the two models in the second stage is as follows:

\begin{promptbox}[Second-stage differential-verification input convention]{gray}
In the second filtering stage (differential verification), the input convention of the Gemini multimodal model and the text-only JUDGER is:

{}[Image description · system]\\
You are an image-description assistant. Describe the key visual information in this Xiaohongshu image (scene/subject/text in the image/objects) objectively in 1\textasciitilde{}2 Chinese sentences to help understand the accompanying text. Output only the description, without evaluation.

{}[Translation · system]\\
You are a professional Chinese-to-English translator for Xiaohongshu. Output only the English translation, without any explanation or quotation marks.

{}[No-image translation · user]   Translate into English: \{text\}\\
{}[With-image translation · user] Please translate into English based on the accompanying image: \{text\}   (+ native image)

{}[JUDGER context]  the image description above is filled into the reference-context slot; both translations are scored under the same context\\
{}[Keep criterion]  gap = score(with-image) - score(no-image);   gap \textgreater{}= 1  =\textgreater{}  image-dependent, keep
\end{promptbox}

\subsection{Privacy-Preserving Rewriting Protocol}
\label{app:rewrite}

All studied phenomena and candidate instances derive from real UGC, but the original posts may contain identifiable people, entities, places and reverse-retrievable visual details, so each is rewritten for privacy before release. The rewrite preserves the vertical, social-media register,
narrative function, evidence-dependency type and core translation point of the original, while replacing people, nicknames, numbers, brands, plot details and surface visual elements that could lead to identification or content traceback. The title, body and image are rewritten jointly, so that the released sample is still an internally consistent, natural UGC note. Each sample uses a
translation-point ledger to record the source expression, target meaning, evidence carrier, clues that must be preserved and shortcuts that must not be introduced, serving as a shared constraint for rewriting and acceptance.

The ledger records five fields per translation point: the source expression (a word, phrase or elided reference in the original), the target meaning (the meaning the English translation must express), the evidence carrier (one of text-only, image, or image-and-text jointly required), the clues that must be preserved (the minimal contextual signal that must survive the rewrite), and the
shortcuts that must not be introduced (any change that would make the answer explicit or easier to produce).

\begin{table}[t]
\caption{Four image-rewriting strategies.}
\label{tab:image_strategy}
\begin{center}
\small
\renewcommand{\arraystretch}{1.3}
\begin{tabular}{@{}p{3.0cm} p{1.9cm} p{6.4cm}@{}}
\toprule
\textbf{Strategy} & \textbf{Mode} & \textbf{Rule} \\
\midrule
{Real Scene Grounded} & image-to-image
& Keep the UI/elements recognizable; change only the composition. Rewrite in-image text
that is a translation point. \\
{Celebrity} & image-to-image
& Preserve the likeness; change viewpoint, pose and composition. Keep the nickname. \\
{Text in Image} & text-to-image
& Change the layout; rewrite in-image text into a semantically equivalent synthetic version. \\
{Generic} & text-to-image
& Rebuild fully with unidentifiable fictional people or objects. \\
\bottomrule
\end{tabular}
\end{center}
\end{table}

Images are rewritten with one of four strategies according to their role in the translation point (Table~\ref{tab:image_strategy}). For real-scene and public-figure images, we preserve the recognizability of the task-relevant subject or scene while changing composition, viewpoint and
non-key details; for in-image text and generic scenes, we regenerate semantically equivalent text or fictional people and objects. All images are generated with gpt-image-2 and pass a perceptual-hash and structural-similarity check, ensuring the released image is sufficiently separated from the source at the visual surface while retaining the evidence needed to complete the translation.

\subsection{Rewriting and Acceptance Constraints}
\label{app:acceptance}

Privacy-preserving rewriting has three steps: generation, independent acceptance and necessary revision. Rewriting is constrained by the translation-point ledger: the cultural expression in \typesym{} must not be explanatorily unpacked, the key meaning of \typevis{} must still depend on
the image, and the cultural and visual evidence of \typehyb{} must both hold; the title, body and image must be updated jointly and remain naturally consistent, while elements that could point to a real identity or the original post are replaced. Each sample also retains its rewriting strategy, evidence-preservation status and privacy-check record to support later re-review.

Text rewriting keeps the original language, social-media register and the implicitness level of each translation point, does not explain the cultural expression in the body, and does not turn a \typevis{} sample into one answerable from text alone. Identifiable identities and private
attributes are replaced with fictional, non-specific content; sentence patterns, narrative angle and characteristic phrases are sufficiently rewritten to reduce surface similarity to the source post while keeping natural UGC expression.

Rewritten samples undergo independent acceptance, focusing on whether the evidence-dependency type is preserved, whether image and text are consistent, whether the content is sufficiently detached from the original post, and whether any privacy or traceable information remains. Failed samples are re-accepted after targeted revision, and are removed from the released set if they still do not meet the requirements.

\subsection{Released-Set Composition and Privacy Checks}
\label{app:release_checks}

The sealing stage performs a rewrite-quality check on all $498$ samples. On the text side, the mean character 3-gram repetition rate is $0.034$ and the mean SequenceMatcher similarity is $0.216$; on the image side, the mean pHash Hamming distance is $30.5$ with a minimum of $12$, and the mean SSIM
is $0.192$. Table~\ref{tab:privacy} summarizes these metrics, showing that the released version has significantly reduced its textual and visual surface similarity to the source posts. The released set consists of $166$ each of \typesym{}, \typevis{} and \typehyb{}, covers $20$ verticals, and contains $518$ images in total.

\begin{table}[t]
\caption{Sealing-stage rewrite-quality metrics (full set of 498 notes).}
\label{tab:privacy}
\begin{center}
\renewcommand{\arraystretch}{1.2}
\begin{tabular}{@{}p{7.2cm} c c c@{}}
\toprule
\textbf{Metric} & \textbf{Mean} & \textbf{Median} & \textbf{Worst case} \\
\midrule
Text 3-gram repetition rate ($\downarrow$)        & 0.034 & 0.024 & max 0.191 \\
SequenceMatcher similarity ($\downarrow$)          & 0.216 & 0.205 & max 0.590 \\
Image pHash Hamming distance\,/\,64 ($\uparrow$)   & 30.5  & ---   & min 12 \\
Image SSIM ($\downarrow$)                          & 0.192 & ---   & max 0.680 \\
\bottomrule
\end{tabular}
\end{center}
\vspace{-0.3\baselineskip}
\end{table}

\section{Human Annotation and Quality Control}
\label{app:annotation}

\subsection{Annotators and Task Flow}
\label{app:annotators}

All data annotation was carried out by three graduate students with an artificial-intelligence background; candidate filtering, type determination, privacy-preserving rewriting and sealing re-review lasted two weeks in total. The annotation interface shows the Chinese image description, the title and body, the corresponding image and the candidate label simultaneously, and annotators
could cross-check against the original note. Each candidate is kept or removed by a human, and grouped into \typesym{}, \typevis{} or \typehyb{} by the evidence the translation requires; model pre-labeling is used only as an aid for filtering. The basis for \typevis{} and \typehyb{} judgments includes: person or object reference requiring the image to resolve, quantity, attribute or semantic ambiguity resolved by the image, or in-image text carrying key meaning. \typesym{} requires that internet-culture knowledge is necessary to complete the translation while the image provides no key semantics.

\subsection{Sealing-Stage Quality Control}
\label{app:sealing_qc}

Sealing re-inspection includes two checks: ``is the image description accurate'' and ``is the privacy-preserving rewrite qualified.'' The image description must cover the subject, scene, attributes or in-image text needed to complete the translation, and must not add information not present in the image; the rewrite check verifies the evidence-dependency type, image-text consistency, naturalness of social-media expression and privacy risk. Failed samples are removed or manually revised and re-accepted.

\subsection{Human Annotation for Judge Validity}
\label{app:judge_annotation}

The judge-validity experiment uses an independent scoring task. We stratified-sample $60$ notes from the benchmark ($20$ each of \typesym{}, \typevis{}, \typehyb{}), and the $10$ model configurations listed in \ref{sec:judge} each generate one translation under text-only and multimodal conditions, giving $60 \times 10 \times 2 = 1{,}200$ translations to score. Annotators view the human-verified image description, the Chinese source and the English translation, and give an overall score by the $0$--$3$ cultural-effectiveness standard defined in this paper; these model configurations match the translator main experiment, but here they only serve to produce translations covering diverse quality and error modes for validating the judge, not to compare translators. After removing invalid pairs with missing translations or incomplete human annotation ($40$ pairs, involving $2$ notes), we keep $1{,}160$ valid human--\judger{} pairs covering $58$ notes (\typesym{} $18$, \typevis{} $20$, \typehyb{} $20$). Scores are further folded into a binary split of effective ($2$--$3$) and ineffective ($0$--$1$) to compute accuracy, Cohen's $\kappa$ and class-level metrics. The full results are in \ref{sec:judge}.

\section{Implementation and Reproduction Details}
\label{app:implementation}

This appendix supplements the translator decoding and judge inference settings. Each model uses the same prompt and decoding configuration under the four translator conditions, changing only the accessible evidence; real images are all input through the model's native image part, without any external OCR or automatic image-captioning module.

\subsection{Translator Decoding and Runtime Settings}
\label{app:translator_decoding}

Except where limited by the interface, the translator sampling temperature is set to $0.7$; GPT-5.5, GPT-5.6 and Kimi-K3 use the interface default temperature. The maximum generation length is $8192$ tokens, relaxed to $32768$ for long samples of GLM-5.3-Flash. Self-hosted models use $8$-way tensor parallelism ($\mathrm{tp}=8$) with a $32768$ maximum context and thinking mode disabled. The four conditions Text Only, Multimodal, Random Image and Human Caption keep the same prompt and decoding settings, changing only the input evidence.

\subsection{Judge Inference Settings}
\label{app:judge_inference}

The \judger{} uses deterministic decoding, with temperature set to $0.0$, top\_p set to $1.0$, a maximum output length of $2048$ tokens, and thinking mode disabled. The judge reads the same Caption Only evidence---the human-verified Chinese image description---under all translator conditions; the specific input organization is given in \ref{sec:protocol} and Appendix~\ref{app:judge_prompt}. The comparison judges in Table~\ref{tab:judge} include the raw Qwen3.5-27B without domain training, Gemini-3.1-pro, Qwen3.5-397B and Qwen3-32B. They also use
temperature $=0.0$ and follow the same input evidence, scoring rubric and output format as the ··\judger{}, to keep the comparison conditions consistent.

\end{document}